\documentclass{article}
\usepackage{ijcai16}
\usepackage{enumitem}
\usepackage{times}
\usepackage{epstopdf}
\usepackage{graphicx}
\usepackage{amsmath}
\usepackage{algorithm}
\usepackage{algpseudocode}
\usepackage{url}
\usepackage{multirow}

\title{Sentence Level Recurrent Topic Model: Letting Topics Speak for Themselves}

\author{Fei Tian\thanks{This work was done when the two authors were visiting Microsoft Research Asia.}\\
University of Science and Technology of China\\
tianfei@mail.ustc.edu.cn
\And
Bin Gao\\
Microsoft Research\\
bingao@microsoft.com
\AND
Di He\\
Microsoft Research\\
dihe@microsoft.com
\And
Tie-Yan Liu\\
Microsoft Research\\
tyliu@microsoft.com
}

\begin{document}
\maketitle

\begin{abstract}
We propose Sentence Level Recurrent Topic Model (SLRTM), a new topic model that assumes the generation of each word within a sentence to depend on both the topic of the sentence and the whole history of its preceding words in the sentence. Different from conventional topic models that largely ignore the sequential order of words or their topic coherence, SLRTM gives full characterization to them by using a Recurrent Neural Networks (RNN) based framework. Experimental results have shown that SLRTM outperforms several strong baselines on various tasks. Furthermore, SLRTM can automatically generate sentences given a topic (i.e., topics to sentences), which is a key technology for real world applications such as personalized short text conversation.
\end{abstract}

\section{Introduction}

Statistic topic models such as Latent Dirichlet Allocation (LDA) and its variants~\cite{landauer1998LSA,hofmann1999probabilistic,blei2003latent,supervisedLDA,onlineLDA} have been proven to be effective in modeling textual documents. In these models, a word token in a document is assumed to be generated by a hidden mixture model, where the hidden variables are the topic indexes for each word and the topic assignments for words are related to document level topic weights. Due to the effectiveness and efficiency in modeling the document generation process, topic models are widely adopted in quite a lot of real world tasks such as sentiment classification~\cite{mei2007topic}, social network analysis~\cite{LDA4Microblog,mei2007topic}, and recommendation systems~\cite{godin2013using}.

Most topic models take the \emph{bag-of-words} assumption, in which every document is treated as an unordered set of words and the word tokens in such a document are sampled independently with each other. The \emph{bag-of-words} assumption brings computational convenience, however, it sacrifices the characterization of sequential properties of words in a document and the topic coherence between words belonging to the same language segment (e.g., sentence). As a result, people have observed many negative examples. Just list one for illustration ~\cite{wallach2006topic}: \emph{the department chair couches offers} and \emph{the chair department offers couches} have very different topics, although they have exactly the same bag of words.

There have been some works trying to solve the aforementioned problems, although still insufficiently. For example, several sentence level topic models~\cite{HTMM,du2010sequentialLDA,wang2011structural} tackle the topic coherence problem by assuming all the words in a sentence to share the same topic (i.e., every sentence has only one topic). In addition, they model the sequential information by assuming the transition between sentence topics to be Markovian. However, words within the same sentence are still exchangeable in these models, and thus the \emph{bag-of-words} assumption still holds within a sentence. For another example, in \cite{yang2015ordering}, the embedding based neural language model~\cite{bengio2003neural,mikolov2013efficient,le2014distributed} and topic model are integrated. They assume the generation of a given word in a sentence to depend on its local context (including its preceding words within a fixed window) as well as the topics of the sentence and document it lies in. However, using a fixed window of preceding words, instead of the whole word stream within a sentence, could only introduce limited sequential dependency. Furthermore, there is no explicit coherence constraints on the word topics and sentence topics, since every word can have its own topics in their model.

We propose Sentence Level Recurrent Topic Model (SLRTM) to tackle the limitations of the aforementioned works. In the new model, we assume the words in the same sentence to share the same topic in order to guarantee topic coherence, and we assume the generation of a word to rely on the whole history in the same sentence in order to fully characterize the sequential dependency. Specifically, for a particular word $w$ within a sentence $s$, we assume its generation depends on two factors: the first is the whole set of its historical words in the sentence and the second is the sentence topic, which we regard as a pseudo word and has its own distributed representations. We use Recurrent Neural Network (RNN)~\cite{mikolov2010recurrent}, such as Long Short Term Memory (LSTM)~\cite{LSTM} or Gated Recurrent Unit (GRU) network~\cite{GRU}, to model such a long term dependency.

With the proposed SLRTM, we can not only model the document generation process more accurately, but also construct new natural sentences that are coherent with a given topic (we call it \emph{topic2sentence}, similar to \emph{image2sentece}\cite{vinyals2015show}). Topic2sentence has its huge potential for many real world tasks. For example, it can serve as the basis of personalized short text conversation system~\cite{shang-lu-li:2015:ACL-IJCNLP,serban2015building}, in which once we detect that the user is interested in certain topics, we can let these topics speak for themselves using SLRTM to improve the user satisfactory.

We have conducted experiments to compare SLRTM with several strong topic model baselines on two tasks: generative model evaluation (i.e. test set perplexity) and document classification. The results on several benchmark datasets quantitatively demonstrate SLRTM's advantages in modeling documents. We further provide some qualitative results on \emph{topic2sentence}, the generated sentences for different topics clearly demonstrate the power of SLRTM in topic-sensitive short text conversations.

\section{Related Work}
One of the most representative topic models is Latent Dirichlet Allocation~\cite{blei2003latent}, in which every word in a document has its topic drawn from document level topic weights.
%Variational inference (VI)~\cite{blei2003latent} and Gibbs sampling based methods~\cite{heinrich2005parameter} have been developed for posterior inference in LDA.
Several variants of LDA have been developed such as hierarchical topic models~\cite{HDPLDA} and supervised topic models~\cite{supervisedLDA}. With the recent development of deep learning, there are also neural network based topic models such as~\cite{hinton2009replicated,docNADE,NNTopic,GaussianLDA}, which use distributed representations of words to improve topic semantics.

Most of the aforementioned works take the \emph{bag-of-words} assumption, which might be too simple according to our discussions in the introduction. That is, it ignores both sequential dependency of words and topic coherence of words.

There are some efforts trying to address the limitations of the \emph{bag-of-words} assumption. For example, in~\cite{griffiths2004HMMLDA}, both semantic (i.e., related with topics) and syntactic properties of words were modeled. After that, a hidden Markov transition model for topics was proposed~\cite{HTMM}, in which all the words in a sentence were regarded as having the same topic. Such a \emph{one sentence, one topic} assumption was also used by some other works, including ~\cite{du2010sequentialLDA,wang2011structural}. Although these works have made some meaningful attempts on topic coherence and sequential dependency across sentences, they have not sufficiently model the sequential dependency of words within a sentence. To address this problem, the authors of ~\cite{yang2015ordering} adopted the neural language model technology ~\cite{bengio2003neural} to enhance topic model. In particular, they assume that every document, sentence, and word have their own topics and the topical information is conveyed by their embedding vectors through a Gaussian Mixture Model (GMM) as a prior. In the GMM distribution, each topic corresponds to a mixture parameterized by the mean vector and covariance matrix of the Gaussian distribution. The embedding vectors sampled from the GMM are further used to generate words in a sentence according to a feedforward neural network. To be specific, the preceding words in a fixed sized window, together with the sentence and document, act as the context to generate the next word by a softmax conditional distribution, in which the context is represented by embedding vectors. While this work has explicitly modeled the sequential dependency of words, it ignores the topic coherence among adjacent words.

Another line of research related to our model is Recurrent Neural Network (RNN), especially some recently developed effective RNN models such as Long Short Term Memory~\cite{LSTM} and Gated Recurrent Unit~\cite{GRU}. These new RNN models characterize long range dependencies for a sequence, and has been widely adopted in sequence modeling tasks such as machine translation~\cite{GRU} and short text conversation~\cite{shang-lu-li:2015:ACL-IJCNLP}. In particular, for language modeling tasks, it has been shown that RNN (and its variants such as LSTM) is much more effective than simple feedforward neural networks with fixed window size~\cite{mikolov2010recurrent} given that it can model dependencies with nearly arbitrary length.

\section{Sentence Level Recurrent Topic Model}
In this section, we describe the proposed Sentence Level Recurrent Topic Model (SLRTM). First of all, we list three important design factors in SLRTM as below.

\begin{itemize}
\item SLRTM takes the \emph{one sentence, one topic} assumption as in~\cite{HTMM,du2010sequentialLDA,wang2011structural}: all words within the same sentence share the same topic. This assumption guarantees the topic coherence within a sentence, and makes topic2sentence possible.
\item To model long range dependencies between words, SLRTM uses RNN (specifically LSTM) with word embedding vectors as input. The purpose is to leverage word embeddings to enhance the semantics of words, as indicated by the previous neural network based topic models~\cite{hinton2009replicated,docNADE,NNTopic,GaussianLDA}.
\item Each topic in SLRTM has its own distributed representation, which is fine-tuned through the training process and is used to generate the whole sentence. Topic representation vector plays a similar role to the source sentence representation in LSTM based machine translation~\cite{GRU} and the image vector output by Convolutional Neural Network in image captioning~\cite{vinyals2015show}.
\end{itemize}

With the three points in mind, let us introduce the detailed generative process of SLRTM, as well as the stochastic variational inference and learning algorithm for SLRTM in the following subsections.

\subsection{The generative process}
\label{generative}
Suppose we have $K$ topics, $|\mathcal{W}|$ words contained in dictionary $\mathcal{W}$, and $M$ documents $D =\{d_1,\cdots,d_M\}$. For any document $d_i, i\in\{1,2,\cdots,M\}$, it is composed of $N_i$ sentences and its $j$th sentence $s_{ij}$ consists of $T_{ij}$ words. Similar to LDA, we assume there is a $K$-dimensional Dirichlet prior distribution $Dir(\alpha)$ for topic mixture weights of each document. With these notations, the generative process for document $d_i$ can be written as below:
{\footnotesize
\begin{enumerate}
\item Sample the multinomial parameter $\theta_i$ from $Dir(\alpha)$;
\item For the $j$th sentence of document $d_i$ $s_{ij}=(y_1,\cdots,y_{T_{ij}})$, $j\in\{1,\cdots, N_i\}$, where $y_t\in \mathcal{W}$ is the $t$th word for $s_{ij}$:
\begin{enumerate}
\item Draw the topic index $k_{ij}$ of this sentence from $\theta_i$;
\item For $t=1,\cdots,T_{ij}$:
\begin{enumerate}
\item Compute LSTM hidden state $\mathbf{h_t}=f(\mathbf{h_{t-1}};\mathbf{y_{t-1}}; \mathbf{k_{ij}})$;
\item $\forall w\in \mathcal{W}$, draw $y_t$ from
\begin{equation}
\label{eqn:Psentence}
P(w|y_{t-1},\cdots,y_1; k_{ij}) \propto g(\mathbf{w'};\mathbf{h_t};\mathbf{y_{t-1}};\mathbf{k_{ij}})
\end{equation}
\end{enumerate}
\end{enumerate}
\end{enumerate}
}

Here we use bold characters to denote the distributed representations for the corresponding items. For example, $\mathbf{y_t}$ and $\mathbf{k_{ij}}$ denote the embeddings for word $y_t$ and topic $k_{ij}$, respectively. $\mathbf{h_0}$ is a zero vector and $y_0$ is a fake \emph{starting} word. Function $f$ is the LSTM unit to generate hidden states, for which we omit the details due to space restrictions. Function $g$ typically takes the following form:
{\footnotesize
\begin{equation}
\label{eqn:softmax}
g(\mathbf{w'};\mathbf{h_t};\mathbf{y_{t-1};\mathbf{k_{ij}}})= \sigma(\mathbf{w'} \cdot (W_1\mathbf{h_t}+W_2\mathbf{y_{t-1}}+W_3\mathbf{k_{ij}} + b)),
\end{equation}
}where $\sigma(x)=1/(1+exp(-x))$, $\mathbf{w'}$ denotes the \emph{output} embedding for word $w$. $W_1,W_2,W_3$ are feedforward weight matrices and $b$ is the bias vector.

Then the probability of observing document $d_i$ can be written as:
{\footnotesize
\begin{equation}
\label{eqn:prob}
\begin{aligned}
P(d_i|\alpha, \Theta)&= \int_{\theta\sim Dir(\alpha)}\prod_{j=1}^{N_i}\sum_{k=1}^K\theta_{ik} P(s_{ij}|k,\Theta)d\theta\\
&=\int_{\theta\sim Dir(\alpha)}\prod_{j=1}^{N_i}\sum_{k=1}^K\theta_{ik}\prod_{t=1}^{T_{ij}}P(y_t|y_{t-1},\cdots,y_1;k)d\theta
\end{aligned}
\end{equation}
}where $P(s_{ij}|k,\Theta)$ is the probability of generating sentence $s_{ij}$ under topic $k$, and it is decomposed through the probability chain rule; $P(y_t|y_{t-1},\cdots,y_1;k)$ is specified in equation (\ref{eqn:Psentence}) and (\ref{eqn:softmax}); $\Theta$ represents all the model parameters, including the distributed representations for all the words and topics, as well as the weight parameters for LSTM.

\begin{figure}
\centering
\includegraphics[width=0.9\columnwidth]{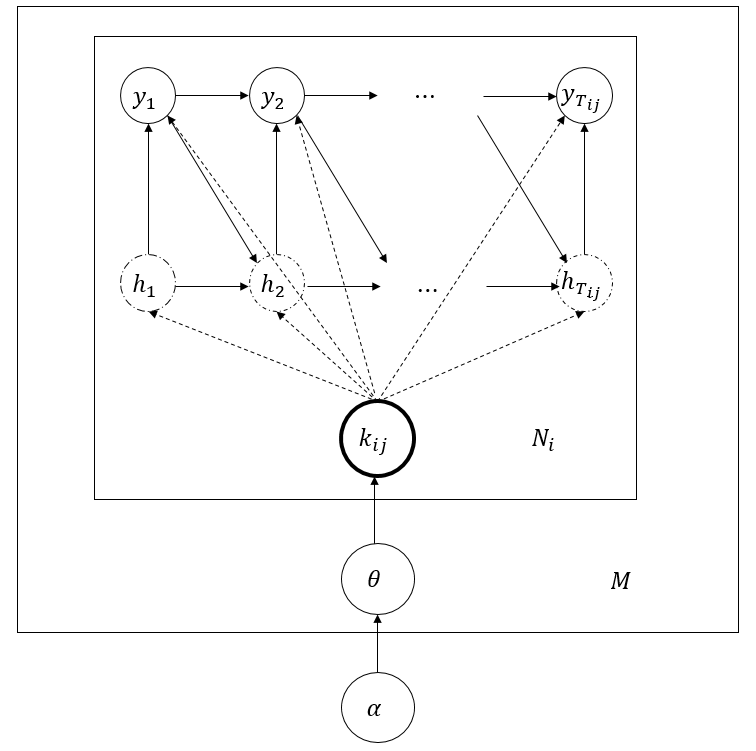}
\caption{The illustration of the SLRTM generative process.}
\label{fig:structure}
\vspace{-3mm}
\end{figure}

To sum up, we use Figure \ref{fig:structure} to illustrate the generative process of SLRTM, from which we can see that in SLRTM, the historical words and topic of the sentence jointly affect the LSTM hidden state and the next word.

\subsection{Stochastic Variational Inference and Learning}
\label{SVI}

As the computation of the true posterior of hidden variables in equation (\ref{eqn:prob}) is untractable, we adopt mean field variational inference to approximate it. Particularly, we use multinomial distribution $q_{\phi_{ij}}(k_{ij})$ and Dirichlet distribution $q_{\gamma_i}(\theta_i)$ as the variational distribution for the hidden variables $k_{ij}$ and $\theta_i$, and we denote the variational parameters for document $d_i$ as $\Phi_i=\{\phi_j, \gamma\},\forall i,j$, with the subscript $i$ omitted. Then the variational lower bound of the data likelihood \cite{blei2003latent} can be written as:

%\footnotesize
\begin{equation}
\label{eqn:ELBO}
\begin{aligned}
\mathcal{L}(D;\Phi,\Theta,\alpha)=&\sum_{i=1}^M\sum_{j=1}^{N_i}\{E_q[\log P(s_{ij}|k_{ij},\Theta)]+E_q[\log \frac{p(k_{ij}|\theta_i)}{q(k_{ij})}]\}\\
&+\sum_{i=1}^ME_q[\log p(\theta_i)-\log q(\theta_i)]
\end{aligned}
\end{equation}
where $p(\cdot)$ is the true distribution for corresponding variables.

The introduction of LSTM-RNN makes the optimization of (\ref{eqn:ELBO}) computationally expensive, since we need to update both the model parameters $\Theta=\{\Theta_1,\cdots,\Theta_D\}$ and variational parameters $\Phi$ after scanning the whole corpus. Considering that mini-batch (containing several sentences) inference and training are necessary to optimize the neural network, we leverage the stochastic variational inference algorithm developed in~\cite{onlineLDA,hoffman2013stochastic} to conduct inference and learning in a variational Expectation-Maximization framework.\footnote{We did not use any recently developed algorithms for inference and learning under deep neural networks such as variational autoencoder \cite{VAE} because they are designed for continuous hidden states while our model includes discrete variables.} The detailed algorithm is given in Algorithm \ref{algEM}. The execution of the whole inference and learning process includes several epochs of iteration over all documents $d_i,i = \{1,2,\cdots,M\}$ with Algorithm \ref{algEM} (starting with $t=0$).

{\footnotesize
\begin{algorithm}[ht]
\caption{Stochastic Variational EM for SLRTM}\label{algEM}
\begin{algorithmic}
\State \textbf{Input}: document $d_i$, variation parameters $\Phi_i^{(t)}$, and model weights $\Theta^{(t)}$.
\For {every sentence minibatch $S=(s_{1},\cdots,s_L)$ in $d_i$}
\State $t=t+1$
\State \textbf{E-Step}:
\Repeat
\For {$l \gets 1,L$}
\State $\forall k\in\{1,\cdots,K\}$, i.e., every topic index:
\State Obtain $\beta_{lk}=\log P(s_l|k,\Theta^{(t-1)})$ by LSTM forward pass.
\State $\phi_{lk}^{(t)}\propto \Psi(\gamma_{k}^{(t-1)})+\beta_{lk}$
%\begin{equation}
\begin{align}
\tilde{\gamma}_{k}&=\alpha + \frac{\sum_{m=1}^MN_m}{L}\sum_l\phi_{lk}^{(t)}\label{eqn:gamma}\\
\gamma_{k}^{(t)} &= (1-\rho_t)\gamma_{k}^{(t-1)} + (1-\rho_t)\tilde{\gamma}_{k}\label{eqn:linear_gamma}
\end{align}
%\end{equation}
\EndFor
\Until{convergence}
\State Collect variational parameters $\Phi_i^{(t)}=\{\phi_{lk}^{(t)}, \gamma^{(t)}\},\forall l,k$.
\State \textbf{M-Step}:
\State Compute the gradient $grad^{(t)}=\frac{\partial \mathcal{L}(S;\Phi^{(t)},\Theta^{(t-1)},\alpha)}{\partial \Theta^{(t-1)}}=\sum_{l=1}^L\sum_{k=1}^K\phi^{(t)}_{lk}\frac{\partial\log P(s_l|k,\Theta^{(t-1)})}{\partial \Theta^{(t-1)}}$
by LSTM backward pass.
\State Use $grad^{(t)}$ to obtain $\Theta^{(t)}$ by stochastic gradient descent methods such as Adagrad~\cite{adagrad}.
\EndFor
\end{algorithmic}
\end{algorithm}
}
In Algorithm \ref{algEM}, $\Psi(x)$ is the digamma function. Equation (\ref{eqn:gamma}) guarantees the estimate of $\gamma_{k}$ is unbiased. In equation (\ref{eqn:linear_gamma}), $\rho_t$ is set as $\rho_t=(\tau_0+t)^{-\kappa}$, where $\tau_0\geq 0, \kappa\in(0.5,1]$, to make sure $\gamma^{(t)}$ will converge~\cite{onlineLDA}. Due to space limit, we omit the derivation details for the updating equations in Algorithm \ref{algEM}, as well as the forward/backward pass details for LSTM ~\cite{LSTM}.

\section{Experiments}
\label{exp}
We report our experimental results in this section. Our experiments include two parts: (1) quantitative experiments, including a generative document evaluation task and a document classification task, on two datasets; (2) qualitative inspection, including the examination of the sentences generated under each topic, in order to test whether SLRTM performs well in the topic2sentence task.

\subsection{Quantitative Results}

We compare SLRTM with several state-of-the-art topic models on two tasks: generative document evaluation and document classification. The former task is to investigate the generation capability of the models, while the latter is to show the representation ability of the models.

We base our experiments on two benchmark datasets:

\begin{itemize}
\item \textbf{20Newsgroup}, which contains 18,845 emails categorized into 20 different topical groups such as \emph{religion}, \emph{politics}, and \emph{sports}. The dataset is originally partitioned into 11,314 training documents and 7,531 test documents\footnote{\url{http://qwone.com/~jason/20Newsgroups/20news-bydate.tar.gz}}.
\item \textbf{Wiki10+}~\cite{wiki10+}\footnote{\url{http://www.zubiaga.org/datasets/Wiki10+/}}, which contains Web documents from Wikipedia, each of which is associated with several tags such as \emph{philosophy}, \emph{software}, and \emph{music}. Following \cite{NNTopic}, we kept the most frequent 25 tags and removed those documents without any of these tags, forming a training set and a test set with 11,164 and 6,161 documents, respectively. The social tags associated with each document are regarded as supervised labels in classification. Wiki10+ contains much more words per document (i.e., 1,704) than 20Newsgroup (i.e., 135).
\end{itemize}

We followed the practice in many previous works and removed infrequent words. After that, the dictionary contains about $32k$ unique words for 20Newsgroup and $41k$ for Wiki10+. We adopted the NLTK sentence tokenizer\footnote{\url{http://www.nltk.org/api/nltk.tokenize.html}} to split the datasets into sentences if sentence boundaries are needed.

The following baselines were used in our experiments:

\begin{itemize}
\item \textbf{LDA}~\cite{blei2003latent}. LDA is the classic topic model, and we used GibbsLDA++\footnote{\url{http://gibbslda.sourceforge.net/}} for its implementation.
\item \textbf{Doc-NADE}~\cite{docNADE}. Doc-NADE is a representative neural network based topic model. We used the open-source code provided by the authors.
\item \textbf{HTMM}~\cite{HTMM}. HTMM models consider the sentence level Markov transitions. Similar to Doc-NADE, the implementation was provided by the authors.
\item \textbf{GMNTM}~\cite{yang2015ordering}. GMNTM considers models the order of words within a sentence by a feedforward neural network. We implemented GMNTM according the descriptions in their papers by our own. \end{itemize}

\subsubsection{Experimental Setting}
For SLRTM, we implemented it in C++ using Eigen\footnote{\url{http://eigen.tuxfamily.org/}} and Intel MKL. For the sake of fairness, similar to~\cite{yang2015ordering}, we set the word embedding size, topic embedding size, and LSTM hidden layer size to be $128$, $128$, and $600$ respectively. In the experiment, we tested the performances of SLRTM and the baselines with respect to different number of topics $K$, i.e., $K=128,256$. In initialization (values of $\Theta^{(0)}$ and $\Phi^{(0)}$), the LSTM weight matrices were initialized as orthogonal matrices, the word/topic embeddings were randomly sampled from the uniform distribution $(-0.015,0.015)$ and are fined-tuned through the training process, $\gamma^{(0)}$ and $\alpha$ were both set to $0.5$. The mini-batch size in Algorithm \ref{algEM} was set as $L=5$, and we ran the E-Step of the algorithm for only one iteration for efficiently consideration, which leads to the final convergence after about 6 epochs for both datasets. Gradient clipping with a clip value of $20$ was used during the optimization of LSTM weights. Asynchronous stochastic gradient descent \cite{ASGD} with Adagrad was used to perform multi-thread parallel training.

\subsubsection{Generative Document Evaluation}

\begin{table}[ht]
\centering
\footnotesize
\caption{Average test perplexity per word of different models. Lower perplexity means better document modelling.}
\label{tbl:perp}
\begin{tabular}{|c|c|c|c|c|}
\hline
\multirow{2}{*}{\begin{tabular}[c]{@{}c@{}}\emph{Dataset}\textbackslash\\ \emph{Model}\end{tabular}} & \multicolumn{2}{c|}{20Newsgroup} & \multicolumn{2}{c|}{Wiki10+} \\ \cline{2-5}
& \multicolumn{1}{l|}{$K=128$} & \multicolumn{1}{l|}{$K=256$} & \multicolumn{1}{l|}{$K=128$} & \multicolumn{1}{l|}{$K=256$} \\ \hline
LDA & 1068 & 944 & 989 & 1007 \\ \hline
Doc-NADE & 966 & 930 & 926 & 884\\ \hline
HTMM & 1013 & 892 & 869 & 927 \\ \hline
GMNTM & 933 & 805 & 790 & 734 \\ \hline
SLRTM & 407 & 395 & 323 & 309 \\ \hline
\end{tabular}
\end{table}

We measure the performances of different topic models according to the perplexity per word on the test set, defined as $perp(D)=\exp \{-\frac{\sum_{i=1}^M\log P(d_i)}{\sum_{i=1}^MN'_i}\}$, where $N'_i$ is the number of words in document $d_i$. The experimental results are summarized in Table \ref{tbl:perp}. Based on the table, we have the following discussions:
\begin{itemize}
\item Our proposed SLRTM consistently outperforms the baseline models by significant margins, showing its outstanding ability in modelling the generative process of documents. In fact, as tested in our further verifications, the perplexity of SLRTM is close to that of standard LSTM language model, with a small gap of about $100$ (higher perplexity) on both datasets which we conjecture is due to the margin between the lower bound in equation (\ref{eqn:ELBO}) and true data likelihood for SLRTM.
\item Models that consider sequential property within sentences (i.e., GMNTM and SLRTM) are generally better than other models, which verifies the importance of words' sequential information.  Furthermore, LSTM-RNN is much better in modelling such a sequential dependency than standard feed-forward networks with fixed words window as input, as verified by the lower perplexity of SLRTM compared with GMNTM.

\end{itemize}

\subsubsection{Document Classification}

In this experiment, we fed the document vectors (e.g., the $\gamma$ values in SLRTM) learnt by different topic models to supervised classifiers, to compare their representation power. For 20Newsgroup, we used the multi-class logistic regression classifier and used accuracy as the evaluation criterion. For Wiki10+, since multiple labels (tags) might be associated with each document, we used logistic regression for each label and the classification result is measured by Micro-$F_1$ score~\cite{rcv1}. For both datasets, we use $10\%$ of the original training set for validation, and the remaining for training.

All the classification results are shown in Table \ref{tbl:classification}. From the table, we can see that SLRTM is the best model under each setting on both datasets. We can further find that the embedding based methods (Doc-NADE, GMNTM and SLRTM) generate better document representations than other models, demonstrating the representative power of neural networks based on distributed representations. In addition, when the training data is larger (i.e., with more sentences per document as Wiki10+), GMNTM generates worse topical information than Doc-NADE while our SLRTM outperforms Doc-NADE, showing that with sufficient data, SLRTM is more effective in topic modeling since topic coherence is further constrained for each sentence.

\begin{table}[ht]
\centering
\footnotesize
\caption{Classification results of different models and different topic numbers $K$. For 20Newsgroup, the measure is accuracy; for Wiki10+, the measure is Micro-$F_1$ score. Higher values mean better classification results.}
\label{tbl:classification}
\begin{tabular}{|c|c|c|c|c|}
\hline
\multirow{2}{*}{\begin{tabular}[c]{@{}c@{}}\emph{Dataset}\textbackslash\\ \emph{Model}\end{tabular}} & \multicolumn{2}{c|}{20Newsgroup} & \multicolumn{2}{c|}{Wiki10+} \\ \cline{2-5}
& \multicolumn{1}{l|}{$K=128$} & \multicolumn{1}{l|}{$K=256$} & \multicolumn{1}{l|}{$K=128$} & \multicolumn{1}{l|}{$K=256$} \\ \hline
LDA & 0.657 & 0.632 & 0.351 & 0.336 \\ \hline
Doc-NADE & 0.670 & 0.646 & 0.462 & 0.471 \\ \hline
HTMM & 0.665 & 0.631 & 0.389 & 0.371 \\ \hline
GMNTM & 0.731 &0.695 & 0.416 & 0.425 \\ \hline
SLRTM & 0.739 &0.722 & 0.483 & 0.489 \\ \hline
\end{tabular}
\end{table}

\subsection{Qualitative Results}

In this subsection, we demonstrate the capability of SLRTM in generating reasonable and understandable sentences given particular topics. In the experiment, we trained a larger SLRTM with 128 topics on a randomly sampled $100k$ Wikipedia documents in the year of 2010\footnote{\url{http://www.psych.ualberta.ca/~westburylab/downloads/westburylab.wikicorp.download.html}} with average $275$ words per document. The dictionary is composed of roughly $50k$ most frequent words including common punctuation marks, with uppercase letters transformed into lowercases. The size of word embedding, topic embedding and RNN hidden layer are set to $512$, $1024$ and $1024$, respectively.

We used two different mechanisms in sentence generating. The first mechanism is random sampling new word $y_t$ at every time step $t$ from the probability distribution defined in equation (\ref{eqn:prob}). The second is dynamic programming based beam search \cite{vinyals2015show}, which seeks to generate sentences by globally maximized likelihood. We set the beam size as $30$. The generating process terminates until a predefined maximum sentence length is reached (set as $25$) or an \emph{EOS} token is met. Such an \emph{EOS} is also appended after every training sentence.

\begin{table*}[ht]
\centering
\footnotesize
\caption{Sentences and words generated under five topics. The first letters of each sentence are changed into capital form.}
\label{tbl:sentence}
\begin{tabular}{|p{12mm}|p{85mm}|p{58mm}|p{18mm}|}
\hline
\textbf{Topic} & \textbf{Sampled Sentences} & \textbf{Top 5 Beam Search Sentences} & \textbf{Words}\\ \hline
Topic 1:\newline Politics & The labour party was a regional political party in south africa. \newline The general election was held in five days later. \newline She lost 10,400 votes in the canada general election for 1974. \newline He was named to chief chairman of serbia. \newline He founded the new cabinet on 29 may 2009.
& He was elected to the parliament. \newline He was a member of parliament.\newline He was elected to the democratic party.\newline He was elected to the legislative assembly.\newline He was a member of the council. & elections \newline liberal\newline parliamentary\newline national\newline party \\ \hline
Topic 2:\newline Movie & The film was written and made for comedy directed by paul. \newline ``King of drama" was the first novel by claude wayne miller. \newline In 1994 , she won the national stars by winning oscar. \newline The film was written and made for comedy directed by paul. \newline Thomas ( born 5 april 1981 ) is an american actor.& The film is based on a 2007 movie.\newline The film is based on a 2006 movie.\newline The film was directed by cameron.\newline The film is released after the 2000 film ``Traffi''.\newline He is well known as well as the film.
& film\newline films\newline director\newline oscar\newline movie\\ \hline
Topic 3: \newline Religion& Joseph henry is the archbishop of vanuatu.\newline Church of st. elizabeth is great.\newline He moved to castle school, where he was curator of the church.\newline In the early nineteenth centuries, the society became a colony of teutonic.\newline He was the patron of saint mary. & He is a member of the church.\newline It was founded by the church.\newline He was a member of roman catholic church.\newline It was founded by the diocese.\newline He is a member of the roman catholic diocese.& bishop\newline saint\newline church\newline st.\newline roman\\ \hline
Topic 4:\newline Military & Seven major paratroopers fight for this battle.\newline Recently, joseph was transferred to the british convoy.\newline List of foreign military journals.\newline In 1980 he started his position at raf college. \newline During world war ii, became a surgeon at the royal irish catholic. &The second world war.\newline It was part of the united states.\newline The first battle of the united states.\newline It was part of the united states army.\newline He was a member of the united states army.& general\newline battle\newline military\newline lieutenant\newline colonel\\ \hline
Topic 5:\newline Location & It is the highest historic place in the 2012 scrapped maps.\newline The lake on the southeast.\newline It has been a high tower in the rim of the guerre river.\newline Secondary temple.\newline It is located approximately in a section of hallway. & It is one of the craters.\newline It is part of the river.\newline It is a part of the crater.\newline It is located on the eastern.\newline It is found in the river.&located\newline lower\newline crater\newline river\newline surface \\ \hline
\end{tabular}
\end{table*}

The generating results are shown in Table \ref{tbl:sentence}. In the table, the sentences generated by random sampling and beam search are shown in the second and the third columns respectively. In the fourth column, we show the most representative words for each topics generated by SLRTM. For this purpose, we constrained the maximum sentence length to 1 in beam search, and removed stop words that are frequently used to start a sentence such as \emph{the}, \emph{he}, and \emph{there}.

From the table we have the following observations:
\begin{itemize}
\item Most of the sentences generated by both mechanisms are natural and semantically correlated with particular topics that are summarized in the first column of the table.
\item The random sampling mechanism usually produces diverse sentences, whereas some grammar errors may happen (e.g., the last sampled sentence for Topic 4; re-ranking the randomly sampled words by a standalone language model might further improve the correctness of the sentence). In contrast, sentences outputted by beam search are \emph{safer} in matching grammar rules, but are not diverse enough. This is consistent with the observations in ~\cite{serban2015building}.
\item In addition to topic2sentece, SLRTM maintains the capability of generating words for topics (shown in the last column of the table), similar to conventional topic models.
\end{itemize}

\section*{Conclusion}

In this paper, we proposed a novel topic model called Sentence Level Recurrent Topic Model (SLRTM), which models the sequential dependency of words and topic coherence within a sentence using Recurrent Neural Networks, and shows superior performance in both predictive document modeling and document classification. In addition, it makes topic2sentence possible, which can benefit many real world tasks such as personalized short text conversation (STC).

In the future, we plan to integrate SLRTM into RNN-based STC systems \cite{shang-lu-li:2015:ACL-IJCNLP} to make the dialogue more topic sensitive. We would also like to conduct large scale SLRTM training on bigger corpus with more topics by specially designed scalable algorithms and computational platforms.

%% The file named.bst is a bibliography style file for BibTeX 0.99c
\bibliographystyle{named}
\bibliography{TSpeak}

\end{document}